# SCORE: Replacing Layer Stacking with Contractive Recurrent Depth

Guillaume Godin[1]

[1]Osmo Labs PBC New York, USA

Corresponding authors: guillaume@osmo.ai

**Abstract**

Residual connections are central to modern deep neural networks, enabling stable optimization and efficient information flow across depth. In this work, we propose SCORE *(Skip-Connection ODE Recurrent Embedding)*, a discrete recurrent alternative to classical layer stacking. Instead of composing multiple independent layers, SCORE iteratively applies a single shared neural block using an ODE (Ordinary Differential Equation) inspired contractive update :

$$h_{t+1} = (1 - \Delta t) * h_t + \Delta t * F_\theta(h_t)$$

This formulation can be interpreted as a depth-by-iteration refinement process, where the step size $\Delta t$ explicitly controls stability and update magnitude. Unlike continuous Neural ODE approaches, SCORE uses a fixed number of discrete iterations and standard backpropagation without requiring ODE solvers or adjoint methods.

We evaluate SCORE across graph neural networks (ESOL molecular solubility), multilayer perceptrons, and Transformer-based language models (nanoGPT). Across architectures, SCORE generally improves convergence speed and often accelerates training.
SCORE is reducing parameter count through shared weights. In practice, simple Euler integration provides the best trade-off between computational cost and performance, while higher-order integrators yield marginal gains at increased compute.

These results suggest that controlled recurrent depth with contractive residual updates offers a lightweight and effective alternative to classical stacking in deep neural networks.

# Introduction

Residual connections are a cornerstone of deep neural networks, enabling stable optimization and efficient information flow across many layers. Additive skip connections have proven effective in vision models such as ResNet and in sequence models. In this work, we propose replacing a stack of layers with recurrent refinement steps through a shared block, by revising the skip connection so that it mimics a discretized Ordinary Differential Equation (ODE). This approach applies to any sequential architecture with identical dimensions; we evaluate it on graph convolutional networks, Transformers, and deep feedforward networks.

Existing ODE-based neural networks convert standard architectures into a continuous ODE and solve it with a dedicated solver; examples include Graph Neural ODE(1) and Neural ODE(2). We bypass the need for a continuous ODE solver by generalizing the skip-connection update in the spirit of a discretized ODE(3). Rather than treating residuals as simple additive shortcuts, we reinterpret the residual term as a velocity field governing embedding evolution, and in GNNs, message passing, under a discretized ODE. We evaluate several numerical integrators (Euler, Heun(3), Midpoint, RK4) and review the impact of the method on GNNs for the molecular solubility benchmark ESOL(4) and on nanoGPT(5,6) with the Shakespeare dataset as well as Autosearch 5 min challenge.

We refer to this approach as SCORE (Skip-Connection ODE Recurrent Embedding): the sequence of layers is replaced by recurrent steps that evolve the embedding according to a discretized ODE (fig 1). Empirically, we generally observe improved convergence stability and faster optimization across multiple architectures. This behavior is also slightly observed for nanoGPT trained on the Shakespeare corpus and autosearch challenge. Simple Euler integration offers the best trade-off between performance and cost; Heun or RK4 can yield slight gains at higher computational cost.

Residual skip connections have become ubiquitous since ResNet(7), where they mitigate vanishing gradients and ease optimization. Sander et al. explored the classical ResNets stacking version with adjoint method (3) as well as the Heun example; they did not use any recurrence layers in their ResNets examples. In graph neural networks, the same additive residual formulation often yields mixed results; beneficial for some architectures (e.g. GAT(8)) but detrimental for others (e.g. MPNN(9), DMPNN(10), Graph Transformers(11)) in our experiments. A limitation of classical stacking is that depth is implemented as the composition of independent transformations, without explicit control over the magnitude or stability of iterative updates. In contrast, a dynamical perspective treats depth as an evolution process governed by controlled update rules. More generally, a continuous-time view allows embeddings to evolve according to a differential equation rather than a fixed discrete update.

To obtain a dynamic, ODE-inspired skip connection without a continuous solver, we adopt a simplified ODE analogy. The Graph Neural ODE(1) was first proposed in 2019 and relies on a continuous ODE formulation. We do not follow that route, as we use a fixed number of discrete steps with a simple Euler-style update (the residual as velocity) and do not introduce any continuous ODE solver or adjoint gradients. The update is a single Euler step per "layer": the embedding is updated by adding a scaled residual (difference term), yielding a lightweight recurrence that can be applied to GNNs, dense networks, and Transformers alike.

Several architectural paradigms exist for deep models with repeated transformations: (i) classical stacking of independent layers with or without residual connections, (ii) parameter tying across depth as in ALBERT-style models(12), and (iii) recurrent depth refinement such as the Universal Transformer(13). SCORE belongs to the third family in that it iteratively applies a single block across steps, but differs in its explicit ODE-motivated contractive update rule (equation 1). Unlike continuous Neural ODE models, SCORE uses a fixed number of discrete iterations and does not rely on an ODE solver or adjoint method. The step size Δt directly controls stability and contraction properties of the update.

Prior work has explored parameter-efficient architectures through tied parameters and iterative refinement. For example, ALBERT shares parameters across layers to reduce model size while maintaining performance. The Universal Transformers introduce a recurrent mechanism across depth to refine representations iteratively using the same transformation function.

In this perspective, stacking corresponds to a sequence of independent operators, while SCORE interprets depth as the repeated application of a single operator under a controlled dynamical update.

Recent work has explored recurrent reasoning models for symbolic tasks (14) (e.g., Sudoku or ARC-AGI). These approaches focus on iterative reasoning rather than architectural depth reduction and are therefore outside the scope of this work.

SCORE can be interpreted as a Krasnosel'skii–Mann-style relaxed fixed-point iteration applied to a learnable operator $F_\theta$, while recurrent reasoning models typically employ the unrelaxed recurrence $h_{t+1} = F_\theta(h_t)$ . Under this view, plain recurrent iteration appears as the special case $\alpha = 1$, and SCORE generalizes it through an explicit relaxation parameter that modulates update stability and dynamics. Empirically, SCORE often performs well with substantially reduced dropout, consistent with an implicit regularization effect induced by shared parameters and the relaxed iterative update.

Our contributions are:

- We introduce a gated residual formulation for the recursive application of a shared neural block.

- Graph neural networks: replacing stacked convolutions with recurrent Euler residual steps and a single shared convolution generally improves convergence stability.

- Dense networks: replacing stacked dense layers with recurrent Euler residual steps and a single shared dense layer maintains performance while reducing parameter count.

- Transformers: replacing stacked decoder blocks with recurrent Euler residual steps using a shared block yields competitive performance on nanoGPT with a smaller number of parameters.

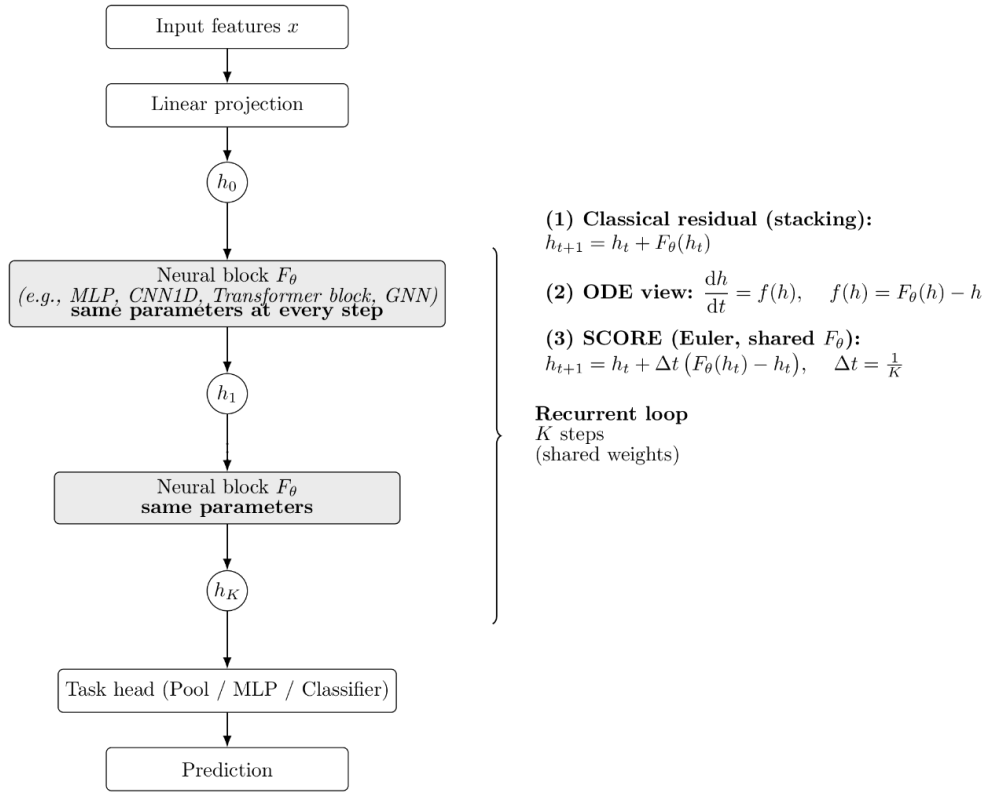

Figure 1: SCORE skip-connection equation using recurrent layer in GNN

# SCORE equation

In contrast to classical stacking of independent layers {$F_1$, $F_2$, …, $F_k$}, SCORE uses a single neural block F whose parameters are shared across steps. The same block is iteratively applied K times, producing a depth-by-iteration refinement process rather than a composition of distinct layers. The residual can be interpreted as a velocity field governing embedding evolution across propagation steps. The SCORE formulation is defined by equation 1.

$$h_{t+1} = h_t + \Delta t *( F_\theta(h_t) - h_t ) \quad \text{(equation 1)}$$

The parameters of $F_\theta$ are tied across all iterations t = 1,...,K, making SCORE a recurrent depth formulation rather than a stacked architecture. It can be rewritten as a weighted contractive residual recurrence equation 2. For example, Δt = 0.5 corresponds to averaging the previous embedding and the transformed embedding.

$$h_{t+1} = (1 - \Delta t) * h_t + \Delta t * F_\theta(h_t) \quad \text{(equation 2)}$$

For Δt in [0,1], this update corresponds to a convex interpolation between the previous embedding and the transformed embedding. The parameter Δt therefore directly controls the magnitude of the update and can induce a contractive behavior when F is Lipschitz-bounded. In practice, this stabilizes the iterated application of the shared block and mitigates divergence or oversmoothing.

We can consider the SCORE as a static residual gate. In our study two Δt were used 0.5 or the inverse of number or recurrent steps both give similar results.

## Stability and Step Size Interpretation

SCORE is derived from a first-order explicit Euler discretization of a differential equation of the form:

$$\frac{dh}{dt} = F_\theta(h) - h \quad \text{(equation 3)}$$

Applying one Euler step with step size Δt yields Equation (1). In this interpretation, Δt plays the role of a time step controlling how far the embedding evolves at each iteration. A natural conservative choice is Δt = 1/K when using K refinement steps, analogous to refining a discretization with smaller steps. However, in practice we observe that a fixed averaging update Δt = 0.5 is equally stable and often slightly more effective. Empirically, both schedules produce stable dynamics across architectures, with Δt acting as a simple and effective stability knob rather than a parameter requiring delicate tuning.

## Euler Simplified family

We explore several numerical integrators to approximate the ODE(15) solution using one of the following methods:

- Euler (equation 1 and 2)
- Heun (RK2)
- Midpoint
- Runge–Kutta 4 (RK4)

Importantly, unlike Neural ODE approaches that rely on adaptive continuous solvers and adjoint backpropagation, SCORE fixes the number of discrete steps K and uses standard backpropagation through the unrolled iterations.

While higher-order methods provide better theoretical accuracy(3), they also increase computational cost due to multiple evaluations of the GNN per layer (see supplementary figures 9 and 10). In default experiments, we use four propagation steps and apply a scaling factor Δt. I decided to define the Δt = 1 / n_steps where n_steps = 4 as default. So in practice, Δt range value is [1/7, 0.5], as we went from 1/2 to 1/7 factors using 2 to 7 steps (see supplementary figures 18-21).

# Experimental Setup

## Dataset

We evaluate SCORE on two tasks: molecular property prediction with graph neural networks and language modeling with Transformers.

**Graph learning.**
We use ESOL as a well-established benchmark dataset for aqueous solubility prediction. We follow a 5-fold cross-validation protocol with an 80/20 train/test split. We report benchmark method using the same CV split as well.

**Language modeling.**
For Transformer experiments, we use the Shakespeare dataset from the Gutenberg project, using the nanoGPT training setup. We use a 90/10 training-validation split. We use the GPT-4o tokenizer, not the character simple split. We used the MLX implementation of nanoGPT as the baseline from Karpathy developments : https://github.com/shakedzy/nanogpt.
Few modifications were tested to modernize the architecture with state of the arts recent progress in the field including Relu2, RMSnorm, RoPE and Normalize Q,K vectors based on the nanoChat https://github.com/karpathy/nanochat, I called this version nanoGPTx. The goal was here to see if we can reduce the number of Transformer layers and keep descend and fast convergence using SCORE.
A second experiment was run with the nanochat MLX version just after the autosearch code was published. In this set up the goal is to get the smaller loss in 5 minutes time. SCORE provides the smaller value with 4 M less parameters than the default version on an Apple MacBook M3 Max 128 Gb computer.

## Graph Neural Network Architectures

We compare native and ODE-residual variants of the following well-known graph neural network architectures. AttentiveFP(16), DMPNN (ChemProp(10))), GAT(8), GATv2(17), GINE, MPNN(9) and Graph Transformer(11). Those models are generally very fast and give good performances especially AttentiveFP and Chemprop.

For each architecture we evaluate five configurations:

- GNN-base: model without skip connections

- GNN-classic: residual connection with LayerNorm

- GNN-skip05: Euler residual averaging ($\Delta t = 0.5$)

- SCORE-GNN: recurrent shared block with Δt = 1/K

- SCORE-GNN-skip05: recurrent shared block with Δt = 0.5

We systematically include the MolAttFP virtual node pooling aggregation instead of the classical pooling for all models by default as in AttentiveFP (see supplementaries for ablation studies).

## Graph neural network training protocol

Models are trained using the Adam optimizer with learning rate of 1e-3 and batch size 32. Training runs for up to 150 epochs per fold to analyse convergence behaviour. All experiments were conducted using the MLX framework. All experiments were done on a M4 Apple Pro version with 24 GB ram memory using a mlx-graphs custom version.

We used recent RIGR features which are tautomer/resonance invariant. All models are plugged into an identical MLP to avoid questioning the MLP final impact of performance between models. It consists of one Dropout 10% followed by 3 layers of respective dimension [128,64,32] using leaky_relu activation function. The final projection is a linear dense output to 1 dimension. The SCORE-MLP is a 128 single layer recurrence using Δt = 1/N in the Euler equation.

The ESOL log10 target was not scaled during the training as it can be the case in literature. So the RMSE root mean squared error is the natural error along the LogSolubility range [−8.057, 1.071]. I use a 32 batch size and a learning rate without any early stopping or learning rate scheduler. I have also investigated the SCORE-MLP using the RDKit 217 features compared to MLP with 4 layers. Random Trees, Boosted Tree, Support vector machine and Lasso linear models were also evaluated using the same dataset in order to compare the performances. I tested CatBoost, XGBoost, LightBoost, Random Forest, SVR and Lasso with feature selection using SHAP values importance from RDKit 217 features. It shows that Catboost with the RDKit 217 features can provide a 0.56 RMSE in CV5 and this is the only method that can have this performance over the 6 methods tested. We did not run any hyperparametrization(18) for the layers dimension.

## Graph neural network molecular features

We optionally augment graph embeddings with a vector of 217 RDKit molecular descriptors.

Special process for extreme and not available numbers in the RDkit matrix:

1. Arcsinh squashing (mask NaN/ Inf)

2. Standard scaling (mask NaN/ Inf)

3. NaN / Inf replaced by zero (ie mean imputation in scaled space)

## nano GPT Experiment

To evaluate the generality of the SCORE formulation, we also apply it to Transformer architectures using nanoGPT.

The main question is whether a single Transformer block can be reused recurrently (SCORE-nanoGPT) instead of stacking multiple blocks. The goal is to evaluate whether recurrent depth improves convergence speed and reduces model size, as this architecture was ultra fine tuned, we do not observe much improvement compared to native models.

We train nanoGPT models with embedding sizes 64 and 384 using the Shakespeare dataset. Models are trained for 10k–15k iterations using Adam or AdamW optimizers.

The model was run for 10000 to 15000 iterations. Two models were tested, Small and Large with respectively 64 or 384 embedding size, with the same context window 32, and 4 different layers or 4 steps with the same layer. I used the GTP-4o tokenizer and start-of-play token as described to be the best settings in the Github experiments. I used the Shakespeare Guntheberg dataset and computed loss to monitor model capabilities, I used Adam or AdamW.

For the nanochat 5 min challenge, I have tested our SCORE recurrent method versus a 0.5 residual connection at every stage (aka skip05). We used 2 different NorMuon implementations with Polar Express approximation and kept the rest of others autosearch (9 March) settings defaults except for PR4 trial.

# Results

## Baseline models using RDKit descriptors

Before analyzing the performance of SCORE on graph neural networks, we first establish reference baselines using classical machine learning models trained on RDKit molecular descriptors. The RDKit feature matrix (217 descriptors) was preprocessed by converting invalid values to NaN, applying arcsinh squashing to limit extreme values, followed by standard scaling and mean-imputation in the scaled space.

We trained several classical machine learning models using the same dataset splits as the neural experiments. Among the tested models, CatBoost achieves the best performance with RMSE = 0.56 ± 0.03 (5-fold cross-validation). This result provides a strong reference baseline for the ESOL dataset. Linear models such as LASSO highlight the intrinsic complexity and non-linearity of the solubility prediction task. Feature selection using SHAP(19) improves linear model performance but still remains below the CatBoost baseline (Table 1).

## Dense networks: MLP vs SCORE-MLP

To verify that the SCORE formulation is not limited to graph architectures, we evaluate its effect on dense neural networks. We compare a classical multilayer perceptron (MLP) with its recurrent counterpart SCORE-MLP, trained using identical data splits and optimization settings for 150 epochs using the Adam optimizer.

The results show that SCORE-MLP achieves similar predictive performance while slightly reducing the variance across folds, indicating that the recurrent formulation stabilizes dense models without degrading accuracy (Figure 1).

**Table 1 — Baseline models using RDKit descriptors**

| Model | RMSE |
| --- | --- |
| **CatBoost**(20) | **0.563±0.03** |
| **XGBoost**(21) | 0.674±0.03 |
| **LightBoost**(22) | *0.614±0.04* |
| **Random Forest**(23) | 0.658±0.06 |
| **SVR**(24) | 0.673±0.03 |
| **Lasso**(25) **(Top-10 features)** | 0.803±0.07 |
| **Lasso (Top-100 features)** | 0.636±0.02 |
| **MLP** | 0.642±0.04 |
| **SCORE-MLP (our method)** | 0.630±0.03 |

**5-fold cross-validation results on the ESOL dataset.**

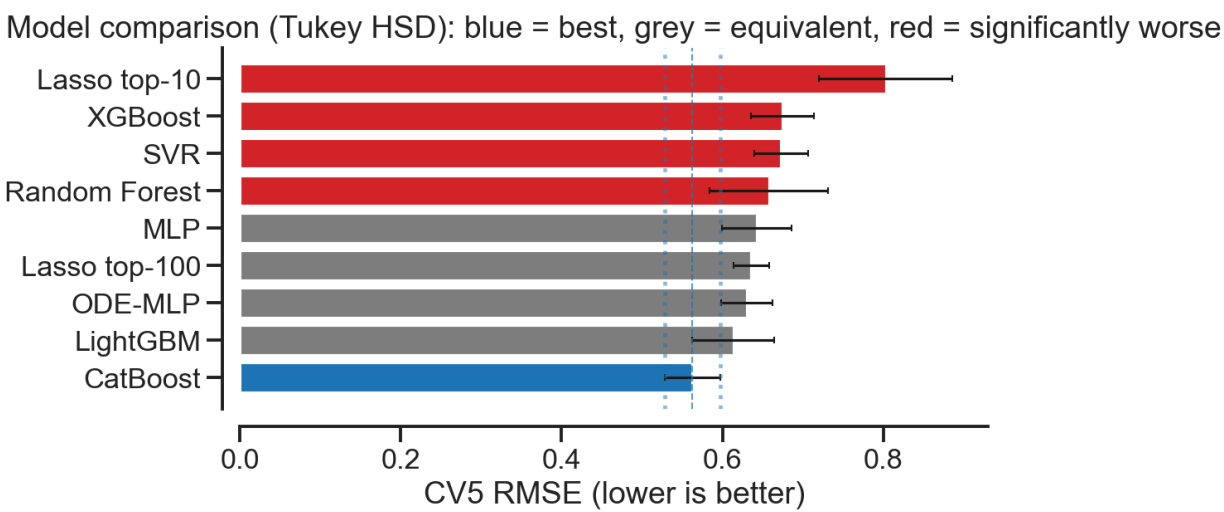

Figure 1 : CV5 benchmarks models RMSE for ESOL prediction lower the better

## Graph Neural Networks

We evaluate the SCORE formulation on a range of graph neural network architectures. To obtain strong GNN baselines, we systematically incorporate the MolAttFP virtual node pooling mechanism, originally introduced in AttentiveFP, across all architectures. This pooling strategy significantly improves the stability of graph models and provides a fair comparison across architectures.

We also include the SCORE-MLP prediction head after graph pooling to maintain a consistent architecture across all models. During training we observed that some architectures such as MPNN and Graph Transformer can be unstable with naive stacking, and benefit from LayerNorm ("classical" residual connections). In contrast, Euler-style skip connections with Δt = 0.5 (skip05) provide stable behavior across most architectures. Overall results show that several SCORE-GNN variants outperform the CatBoost baseline, including DMPNN, AttentiveFP, GINE, GCN, GAT and GATv2. Interestingly, the simple GCN architecture also achieves strong results, demonstrating that the SCORE formulation can effectively propagate embeddings even with lightweight convolution operators.

Across the top-13 performing models (Table 2):

- 10 out of 13 models are SCORE variants

- the second-best performing approach corresponds to the skip05 Euler residual formulation

- both SCORE and skip05 demonstrate strong compatibility with a wide range of GNN architectures

These observations suggest that Euler-style residual updates with controlled step size are well tolerated across graph convolution operators (Figure 2).

Table 2 - Best performing GNN models (5-fold CV)

| Rank | Model | Mean best val RMSE |
|---|---|---|
| 1 | dmpnn_skip05 | 0.533±0.04 |
| 2 | SCORE_dmpnn_skip05 | 0.542±0.05 |
| 3 | SCORE_gat_skip05 | 0.546±0.04 |
| 4 | SCORE_gine | 0.547±0.05 |
| 5 | SCORE_mpnn | 0.555±0.05 |
| 6 | gcn_skip05 | 0.557±0.03 |
| 7 | SCORE_dmpnn | 0.558±0.04 |
| 8 | SCORE_gcn_skip05 | 0.559±0.01 |
| 9 | SCORE_gatv2_skip05 | 0.559±0.03 |
| 10 | gat_skip05 | 0.561±0.04 |
| 11 | SCORE_gine_skip05 | 0.562±0.04 |
| 12 | SCORE_gcn | 0.562±0.03 |
| 13 | SCORE_gat | 0.564±0.04 |

**5-fold cross-validation results on the ESOL dataset.**

Figure 2 : CV5 benchmarks models RMSE for ESOL prediction lower the better

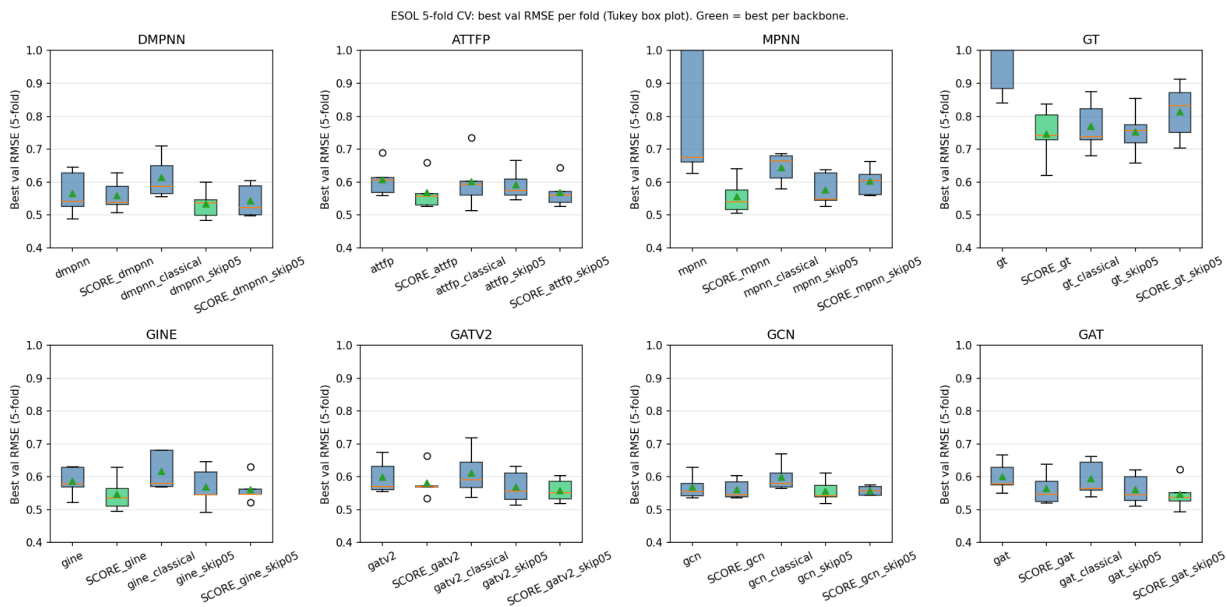

Comparison of the 5 five configurations of 8 GNN architectures

# nanoGPT Experiments

We apply the next SCORE into Transformer models using nanoGPT. We train nanoGPT models on the Shakespeare dataset with embedding dimensions of 64 and 384. Models are trained for 10k–15k iterations using Adam or AdamW optimizers.

Using a larger embedding dimension (384), the SCORE model reaches validation loss 5.41, compared with 5.67 for the native nanoGPT model, despite using fewer parameters (28M vs 34M) see figure 3.

Figure 3 : Train and Validation Loss of nanoGPT variant

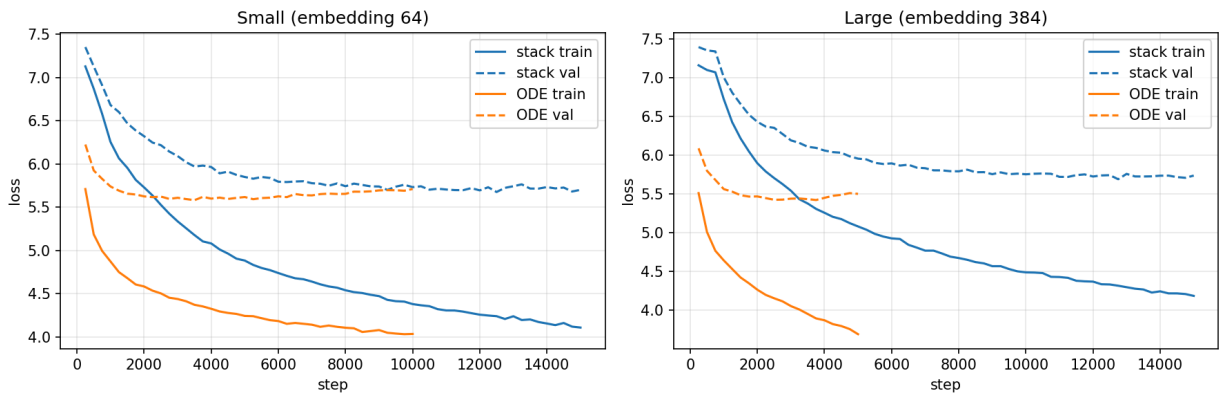

Left small embedding versus right large embedding. SCORE models are learning faster with GTP-4o vocabulary embedding

We also evaluate the modified nanoGPTx architecture with embedding size 64 and varying numbers of recurrent steps. Across experiments, SCORE-based models converge slightly faster and achieve comparable or slightly improved validation loss.

In these experiments, a fixed step size Δt = 0.5 performs slightly better than the theoretical Δt = 1/N schedule, consistent with observations from the GNN experiments (figure 4).

Considering karpathy's autosearch trials, the best setting without Agent intervention, was to use two steps SCORE unique blocks twice so replace d4 by two s2 (SCORE 2 steps) stacked. We got a val_bpb 1.302 after 5 min, 1.282 after 6 min for 18 M parameters. The 4 stacks layers with a skip05 residual (i.e. average) gave val_bpb 1.303 and 1.286 respectively with 22 M parameters. Also by removing the average *skip05* we get the native 4 stacks layers (d4) that gave val_bpb 1.309. Again, the skip05 is improving the native model while the SCORE allows to reduce the parameter number. For references, the H100 Nvidia GPU card for 5 mins gives val_bpb 0.998, as the GPU clock is faster than MPS. One best model (aka 11 March 2026) obtained 1.2809 using a more sophisticated variant of norMuon implementation in d4 after hyperparameter fine tuning using autosearch 110 trials. The major differences to our initial NorMuon was stability, also the batch size of 16 to 8 to allow more iterations in 5 mins.

We were able to reach the val_bpb 1.2594 using our M3 max 128 GB hardware using the skip05 option in the d4 22 M parameters model. The 2 sequential recursive SCORE blocks gave val_bpb 1.2731 which is expected to be worse as the d4 method was fine tuned over 110 trials and because we reduce the parameters to 18.4M instead of 22 M of the original d4. The d4 original model got val_bpb 1.2621 without skip05 (see table 12).

The code is available here : https://github.com/guillaume-osmo/autosearch-mlx.

Figure 4 : Train and Validation Loss of nanoGPT variant

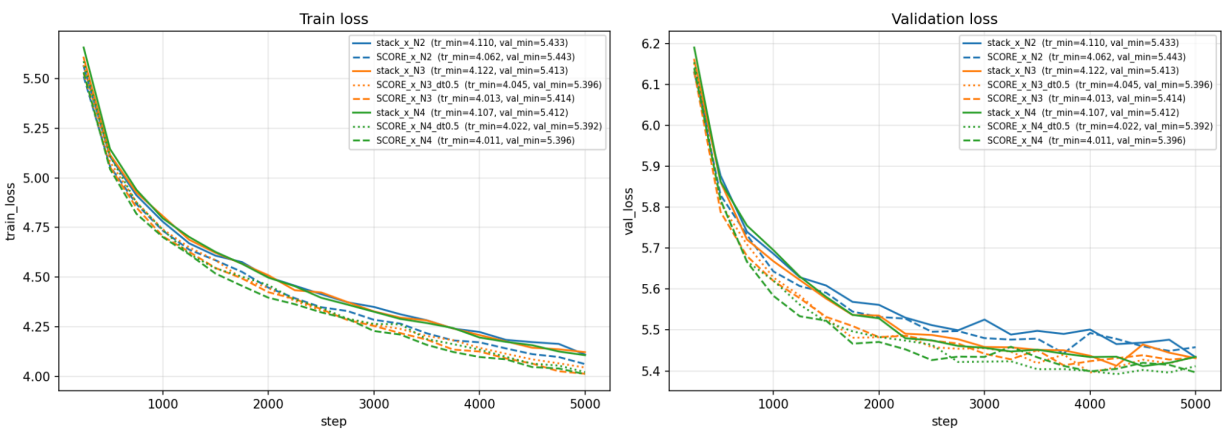

increase the depth of the nanoGPT structure, there is a little improvement for Δt = 1/N versus the 0.5 option

# Discussion

Based on the Lottery ticket assumption(26), only a few portions of weights are really useful in deep learning layers. If we use only one layer initialization we reduce the optimization dimensionality too. We empirically show that a single shared block can effectively support multi-step representation refinement without performance degradation across several architectures. This method generally converges faster and provides better performances. While the established idea that Δt = 1/N is the best in theory, we have seen in our experiments that Δt = 0.5 is generally identical or even better making a good alternative.

NanoGPTx is shown to support ablation of one Transformer step, without similar loss and convergence speed, and a notable parameter reduction. This is very important in the context of LLM size.

Indeed, these results suggest that SCORE provides a more stable and principled mechanism for deep and multi-step message passing. By explicitly modeling the change in embeddings rather than summing representations, SCORE reduces oversmoothing, improves convergence speed and stabilizes training across heterogeneous GNN while being similar for MLP and Transformer already fine tuned architectures.

As we reuse the same convolution weights, the models are smaller but not faster as we do not change the number of steps. The reduction in parameter count may contribute to improved optimization stability by reducing the dimensionality of the parameter space.

Because SCORE reuses a single shared block, models contain fewer parameters. Despite this constraint, we obtain performance comparable to stacked architectures, suggesting that recurrent depth can effectively replace multiple independent layers.

In small-data settings, such as ESOL (~1000 molecules), we observe a more pronounced reduction in training time, whereas in larger-data settings, such as the Shakespeare Gutenberg corpus, the gain is more moderate. This suggests that SCORE may act as an implicit regularizer whose benefits are stronger in low-data regimes. This view is consistent with previous work showing that Graph Transformers benefit from larger multitask datasets and auxiliary targets(27). On ESOL, by contrast, the low-data regime appears to limit Graph Transformer performance, and SCORE partially mitigates this limitation.

The fact that models without SCORE could outperform SCORE variants should be expected, given that SCORE reduces the number of trainable parameters through weight sharing. In our experiments, however, we also observed that this reduction in parameter count can sometimes be beneficial for training, likely by improving optimization stability and acting as an implicit regularizer.

SCORE introduces an implicit iterative refinement loop within each forward pass, which may reduce representational variance similarly to how ensemble averaging or repeated reasoning improves output stability(28).

# Conclusion

We introduce a recursive skip-connection block called SCORE that can be used in MLP, Graph Neural Convolution and Transformers. SCORE goals is to use the same layer recurrently. It is a lightweight yet effective alternative to classical skip connections of multiple layers. Across multiple Graph neural network architectures on ESOL target, the simple SCORE, particularly Euler, with small step size factor or fixed step factor, delivers generally robust improvements in stability and performance without RDKit features. Similarly the use of SCORE-MLP and SCORE-Transformer which maintain competitive convergence and speed.

This work demonstrates that continuous-time reasoning can meaningfully simplify and improve neural network design, without requiring full ODE during training or any adjoint methods.

The Δt can be a learnable parameter of the model per convolution layer : a single trial was done that did not provide better results. A more complete analysis can be done to determine if the step dependent Δt provides nicer results. We can already see that 0.5 or 1/N factors work well.

This is also the first time that we systematically applied the MolAttFP virtual node trick by default in AttentiveFP into all our Graph architectures to get better results without the RDKit features than with the RDKit features. Showing that the generative graph embedding is more efficient than RDKit features. Our GNN alternative obtains better results than CatBoost with SCORE which is generally considered to be part of the best models.

In terms of perspective, we can rethink the needs of several independent layers in Deep learning models. One option that was working is to use several SCORE blocks sequentially, as observed on the nanoGTP2 autosearch example. It would be interesting to leverage it in larger language models. Our skip05 can already stabilize the residual connection even without SCORE blocks.

# Funding

The author was funded by Osmo Labs PBC for Graph Neural Network methods development.

# Competing Interests and Consent for publication

The author declares that he has no competing interests. The author has read and agreed to the published version of the manuscript.

# Acknowledgement

The author wants to thank Brian Kelley and Gregory Landrum for RDKit "217" descriptor c++ implementation support.

# SUPPLEMENTARY

## Ablation of MolAttFP

After removing in all the models (including AttFP) the MolAttFP, we clearly see an average increase of 0.03 RMSE compared to the models including MolAttFP. Again skip 0.5 (average Euler) and SCORE methods generally provide the best models. Only one model is equal to the CatBoost results, showing the very important MolAttFP contribution to GNN in general.

Figure 6 : CV5 distribution comparing the 5 GNN options without MolAttFP

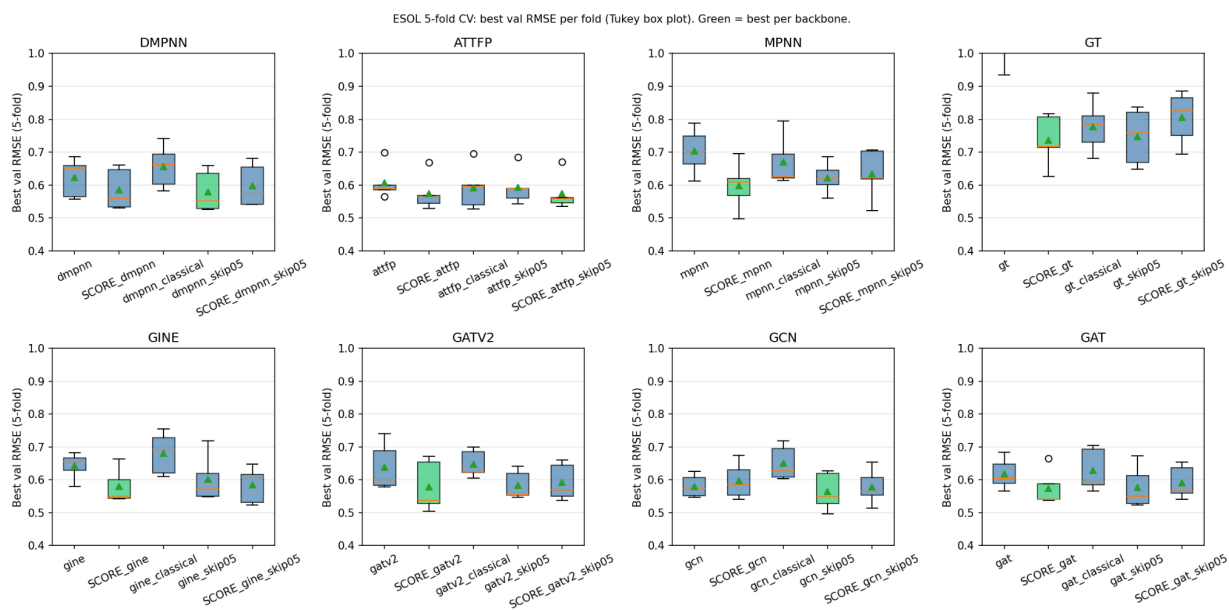

The lower the better in green the best option for this architecture.

Table 3 : Ablation MolAttFP in GNN models.

| Rank | Model | Mean best val RMSE |
|---|---|---|
| 1 | gcn_skip05 | 0.564±0.052 |
| 2 | SCORE_attfp_skip05 | 0.574±0.049 |
| 3 | SCORE_attfp | 0.574±0.049 |
| 4 | SCORE_gat | 0.574±0.049 |
| 5 | gat_skip05 | 0.578±0.057 |
| 6 | SCORE_gcn_skip05 | 0.579±0.048 |
| 7 | SCORE_gatv2 | 0.579±0.070 |
| 8 | dmpnn_skip05 | 0.580±0.056 |
| 9 | SCORE_gine | 0.580±0.047 |
| 10 | gcn | 0.580±0.031 |
| 11 | gatv2_skip05 | 0.583±0.040 |
| 12 | SCORE_gine_skip05 | 0.585±0.049 |
| 13 | SCORE_dmpnn | 0.586±0.057 |

# Ablation of SCORE-MLP

Figure 7 : CV5 distribution comparing the 5 GNN options without SCORE-MLP

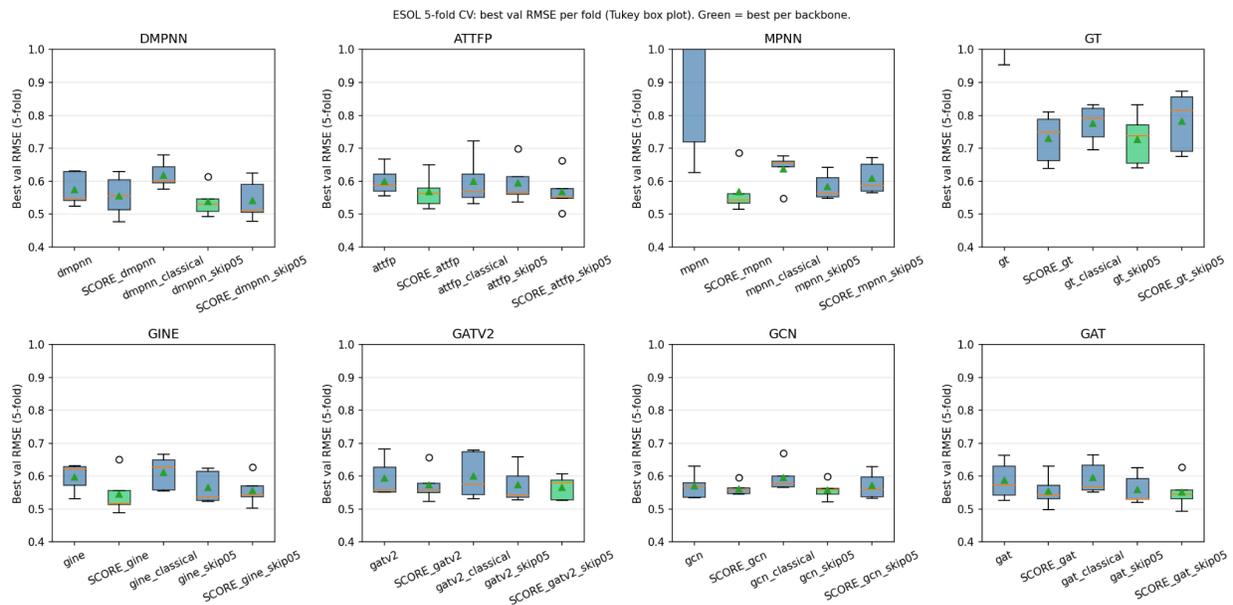

The lower the better in green the best option for this architecture.

Table 4 : Ablation SCORE-MLP in GNN models.

| Rank | Model | Mean best val RMSE |
|---|---|---|
| 1 | dmpnn_skip05 | 0.538±0.042 |
| 2 | SCORE_dmpnn_skip05 | 0.541±0.056 |
| 3 | SCORE_gine | 0.546±0.057 |

| 4 | SCORE_gat_skip05 | 0.551±0.044 |
| 5 | SCORE_dmpnn | 0.555±0.056 |
| 6 | SCORE_gat | 0.555±0.045 |
| 7 | SCORE_gine_skip05 | 0.556±0.042 |
| 8 | gcn_skip05 | 0.557±0.025 |
| 9 | gat_skip05 | 0.560±0.042 |
| 10 | SCORE_gcn | 0.562±0.018 |
| 11 | gine_skip05 | 0.566±0.044 |
| 12 | SCORE_gatv2_skip05 | 0.566±0.033 |
| 13 | SCORE_mpnn | 0.567±0.061 |

## Ablation of MolAttFP and SCORE-MLP

After removing both MolAttFP and SCORE-MLP, we focus on the true effect of SCORE-GNN in the model performance. We got very similar results as for Ablation MolAttFP in general. This means our SCORE-MLP is accepted and does not degrade the performances.

Figure 8 : CV5 distribution comparing the 5 GNN options without both SCORE-MLP and MolAttFP

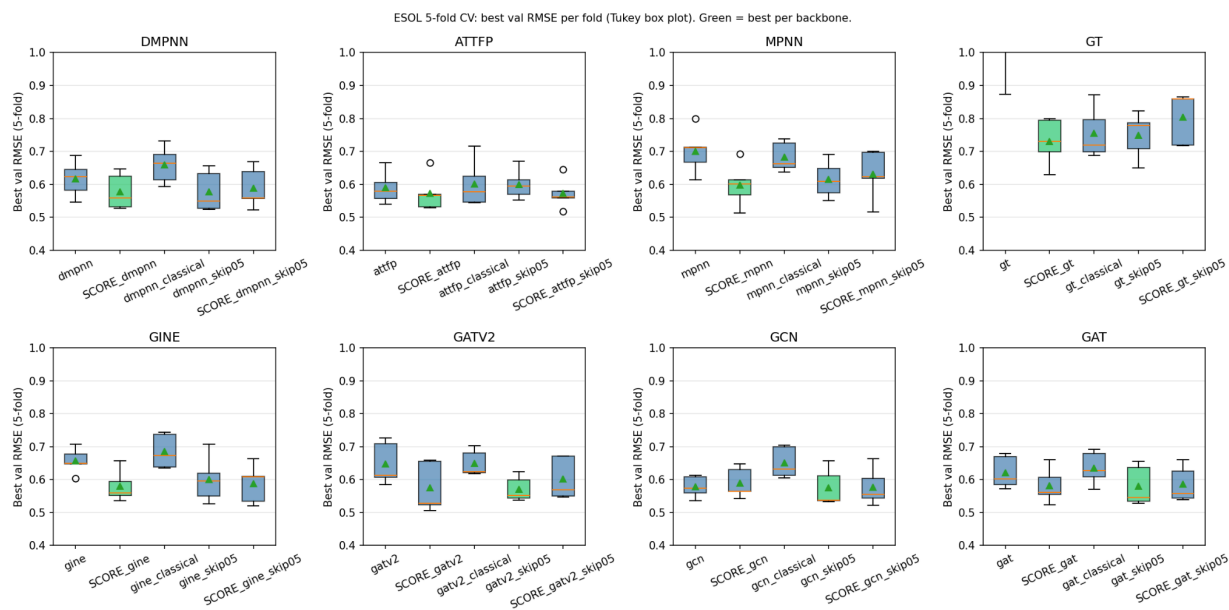

The lower the better in green the best option for this architecture.

Table 5 : Ablation MolAttFP and SCORE-MLP in GNN models.

| Rank | Model | Mean best val RMSE |
|---|---|---|
| 1 | gatv2_skip05 | 0.571±0.034 |
| 2 | SCORE_attfp | 0.572±0.050 |
| 3 | SCORE_attfp_skip05 | 0.573±0.042 |
| 4 | SCORE_gatv2 | 0.575±0.068 |
| 5 | gcn_skip05 | 0.575±0.050 |
| 6 | SCORE_dmpnn | 0.577±0.049 |
| 7 | dmpnn_skip05 | 0.577±0.055 |
| 8 | SCORE_gcn_skip05 | 0.578±0.050 |
| 9 | gcn | 0.578±0.030 |
| 10 | SCORE_gine | 0.579±0.044 |
| 11 | gat_skip05 | 0.580±0.055 |
| 12 | SCORE_gat | 0.581±0.048 |
| 13 | SCORE_gat_skip05 | 0.586±0.049 |

## Study of SCORE equation effect

We run for 75 epochs the models to compare the 4 equations using the same 1/N for 4 steps. we do not see a huge difference versus the complexity of the computation so we keep Euler on the experiments.

Figure 9 : SCORE variations using Euler Simplified family with RDkit features at 75 epochs

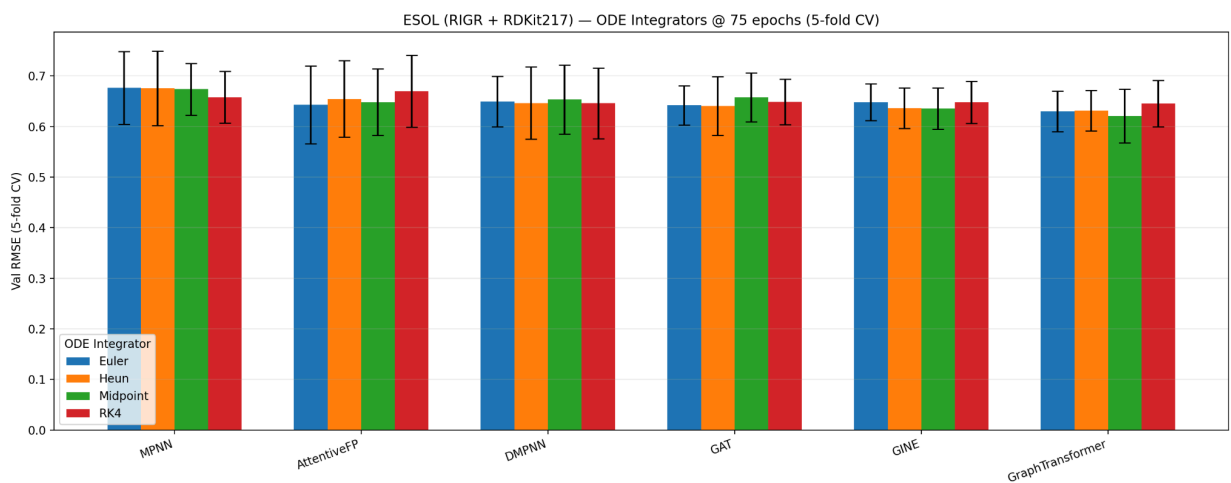

Effect of using more Euler Simplified family (using Euler SCORE equation with 4 steps delta 1/4), without MolAttFP option.

Figure 10 : SCORE variations using Euler Simplified family without RDkit features at 75 epochs

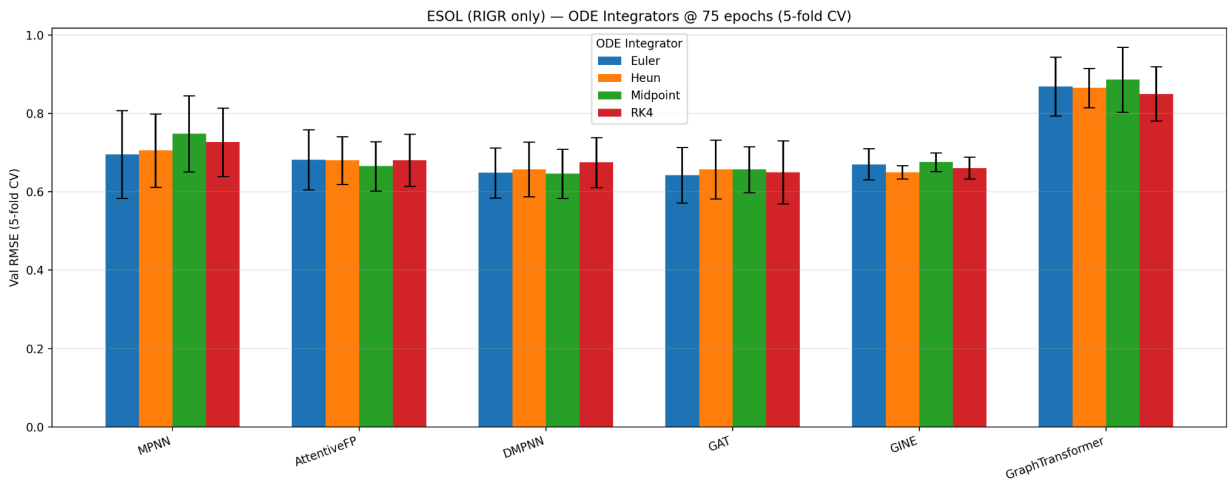

Effect of using more Euler Simplified family (using Euler SCORE equation with 4 steps delta 1/4), without MolAttFP option.

## 2 to 7 convolution layers/steps

In this trial, we investigate the RDKit additional effect using concatenation of 217 descriptors. We generally observed a faster convergence. This is particularly the case for Graph Transformer. Few methods generally outperform or reach similar performance without RDKit. This is the reason we did not include RDKit features in GNNs main studies.

Figure 11 : Validation RMSE of first 100 epochs of SCORE-AttFP vs AttFP

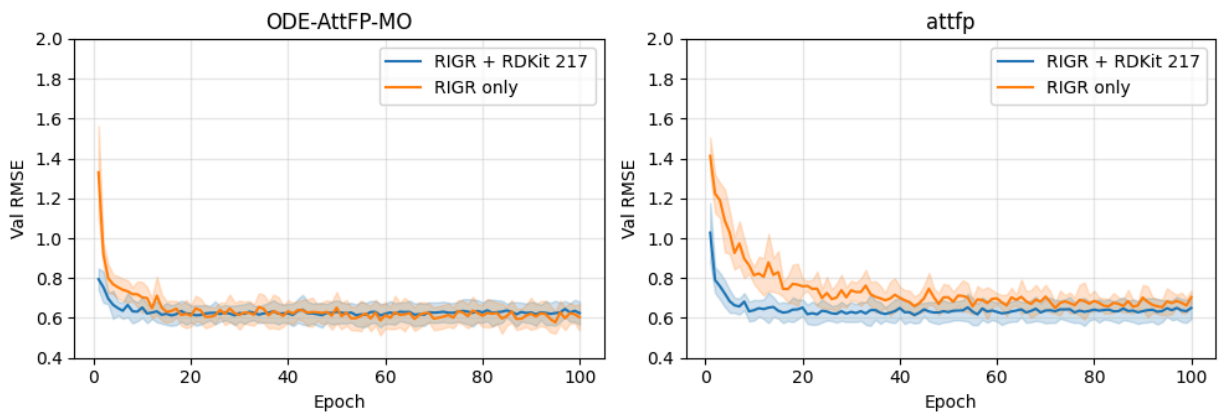

AttFP-MO corresponds to the native AttentiveFP model   validation 5-CV per epoch (using Euler SCORE equation with 4 *steps delta 1/4*).

Figure 12 : Validation RMSE of first 100 epochs of SCORE-GT vs GT

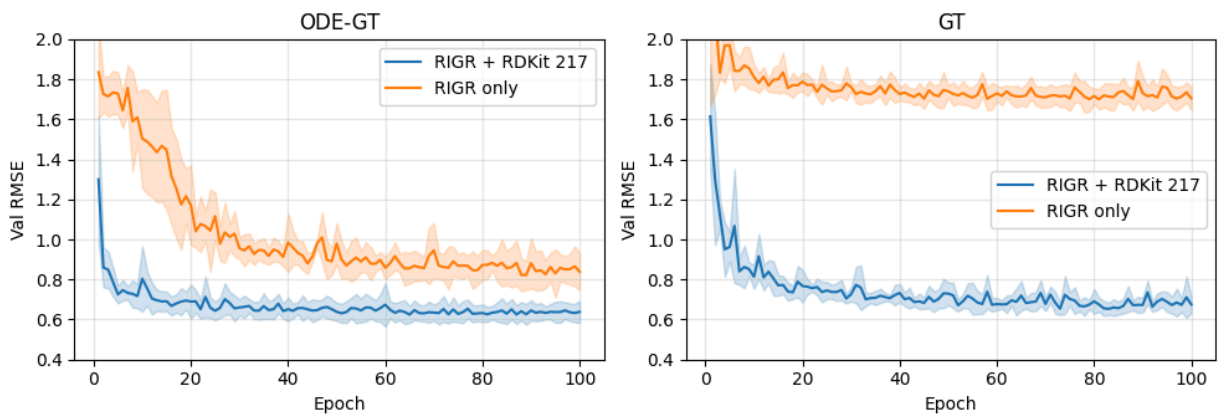

a validation 5-CV per epoch (using Euler SCORE equation with 4 *steps delta 1/4*).

Figure 13 : Validation RMSE of first 100 epochs of SCORE-GAT vs GAT

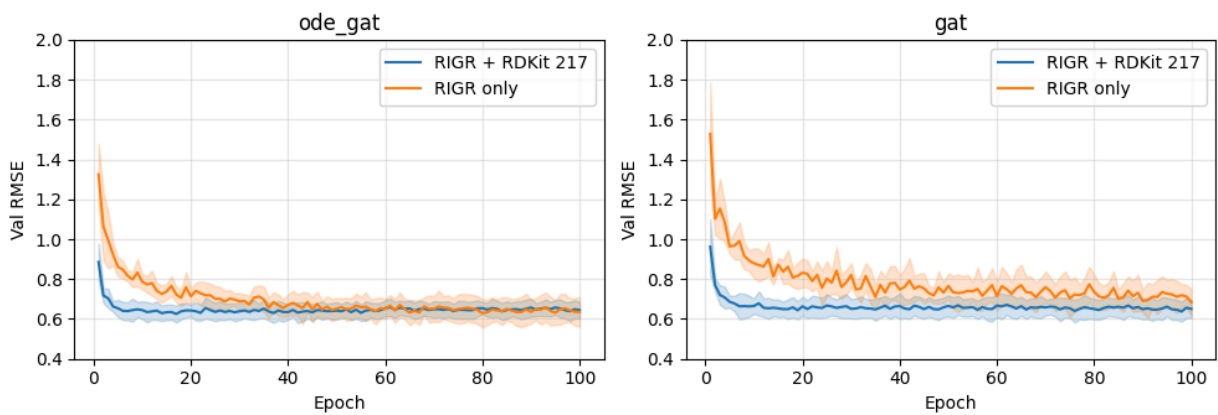

a validation 5-CV per epoch (using Euler SCORE equation with 4 *steps delta 1/4*), 100 epochs.

Figure 14 : Validation RMSE of first 100 epochs of SCORE-GINE vs GINE

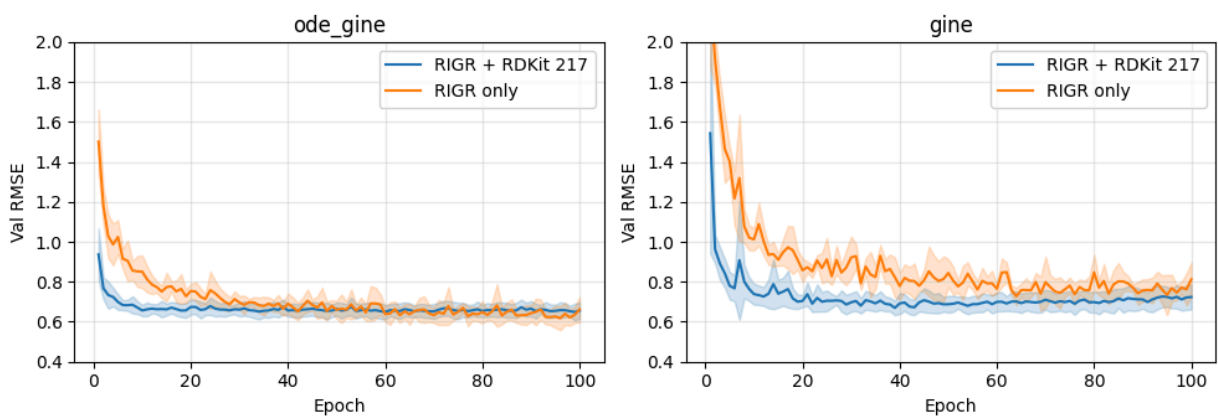

a validation 5-CV per epoch (using Euler SCORE equation with 4 *steps delta 1/4*), 100 epochs.

Figure 15 :  Validation RMSE of first 100 epochs of SCORE-MPNN vs MPNN

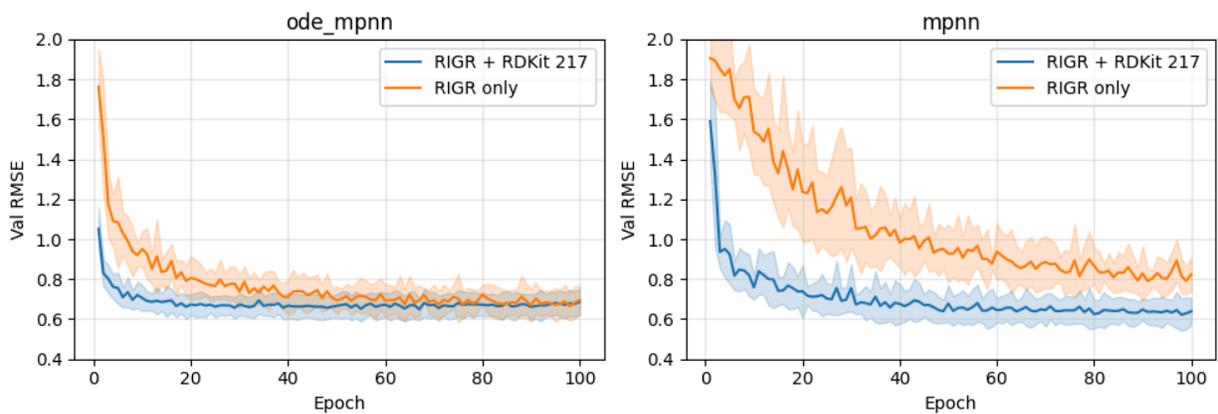

a validation 5-CV per epoch (using Euler SCORE equation with 4 *steps delta 1/4*), 100 epochs.

Figure 16 : Validation RMSE of first 100 epochs of SCORE-DMPNN vs DMPNN

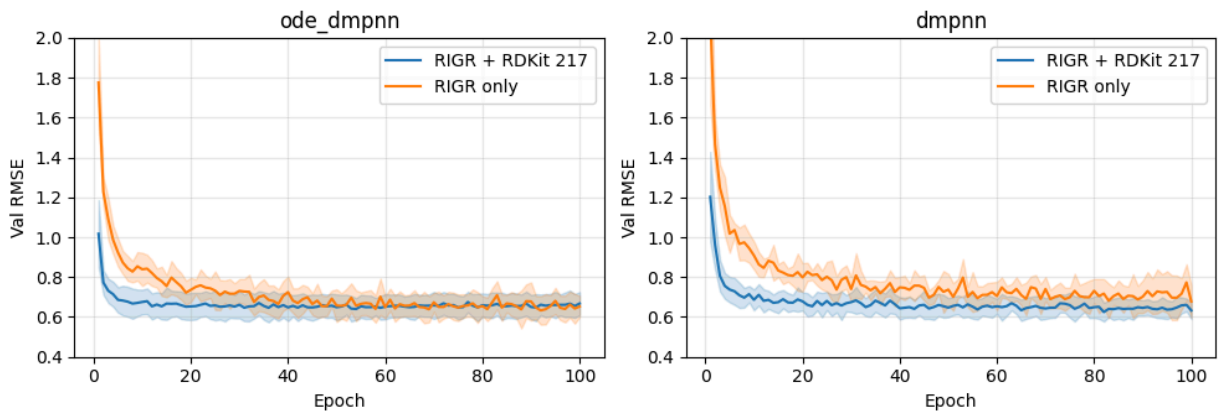

a validation 5-CV per epoch (using Euler SCORE equation with 4 *steps delta 1/4*), 100 epochs.

## SCORE Acceleration versus Native validation

During the experiments, I found that we can map the two learning curves between native and SCORE version initial methods using the time (epoch) - warp alignment of learning curves. The curves have a similar trend with a speed rating difference. This could be useful for two main reasons: find the ideal number of epochs of native methods and make hyperparameter optimization on the SCORE space. For almost all cases except for GT without RDKit, we can fit the two curves using a compression factor on Native curve in order to compute the speed acceleration factor.

Table 6 : Acceleration factor convergence of GNN.

| Speed Acceleration Factor | with RDKit | without RDKit |
|---|---|---|
| **AttFP (with MolAttFP)** | 1.9 | 2.9 |
| **DMPNN** | 1.7 | 1.6 |
| **MPNN** | 2.6 | 6.1 |
| **GINE** | 2.2 | 3.2 |
| **GAT** | 1.5 | 3.3 |
| **GT** | 2.4 | 9.7 |

Factor acceleration of SCORE versus the Native version using RDKit or no, without MolAttFP except AttFP.

Figure 17 : validation loss time warping between SCORE-GAT and GAT

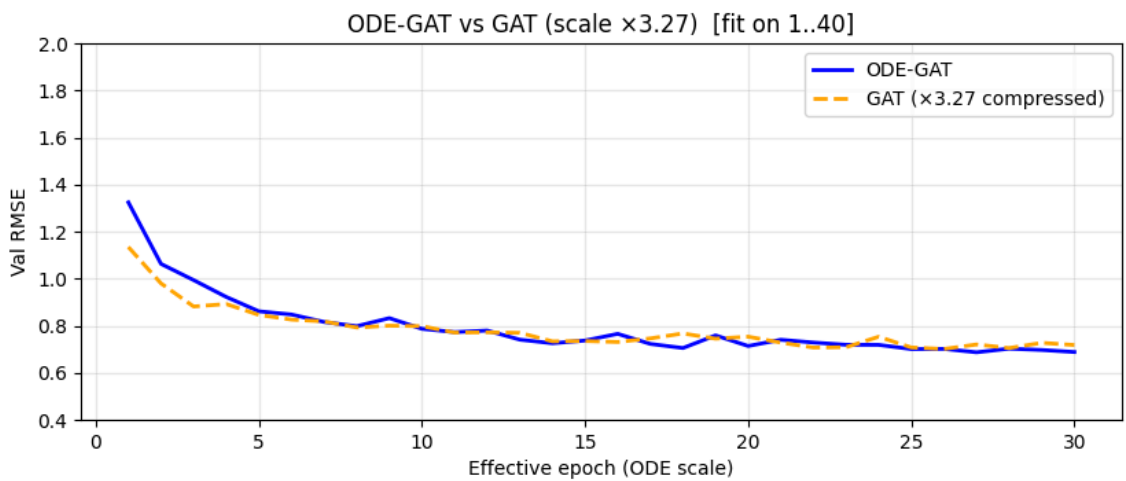

*Example of time-warping fitting between the two methods by compressing the native validation curve fitting.*

This implies that we have a clear speed improvement without losing precision of the model via SCORE method. Interestingly it also provides knowledge of the capabilities and SCORE method versus the original method.

## Oversmoothing analyses

### Study of number of steps / layers in GAT N in [2,7]

Figure 18 : Number of Steps for SCORE-GAT versus GAT without MolAttFP option, dt = 1/N

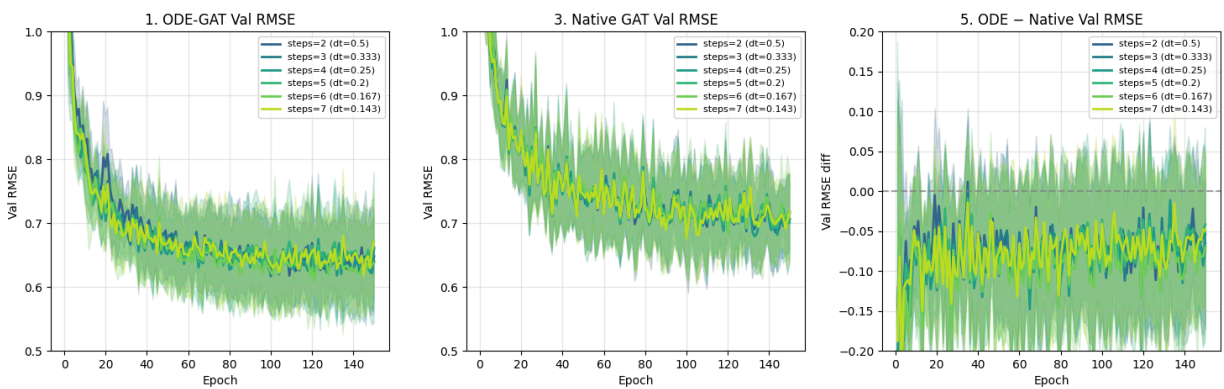

Validation RMSE comparison, by changing number of steps (dt = 1/N) from 2 to 7

In order to understand the robustness of the method to oversmoothing, we run 6 model versions changing the number of steps/layers in GAT structure. I found the same acceleration and improvement performances over the runs with a between 7.6% improvement between the Original GAT and the SCORE-GAT using the best average performance. This is confirming the efficiency of this SCORE method as well as its stability.

Table 7 : SCORE-GAT vs GAT at best validation RMSE

| steps | Δt | SCORE | Native | Diff | Improvement |
|---|---|---|---|---|---|
| 2 | 0.5 | **0.598** | **0.644** | +0.045 | 7.0% |
| 3 | 0.334 | **0.595** | **0.646** | +0.052 | 8.0% |
| 4 | 0.25 | **0.595** | **0.638** | +0.043 | 6.7% |
| 5 | 0.2 | **0.595** | **0.647** | +0.051 | 7.9% |
| 6 | 0.167 | **0.590** | **0.641** | +0.051 | 7.9% |
| 7 | 0.143 | **0.594** | **0.646** | +0.052 | 8.0% |
| **Average** | N/A | **0.595** | **0.644** | +0.049 | **7.6%** |

5-fold CV average of best Validation RMSE over 150 epochs no MolAttFP no SCORE-MLP

## Study of number of steps / layers in GAT + MolAttFP 2 to 7

When starting the experiments I did not use the MolAttFP layer in AttentiveFP to study only the atom graph convolution effect. The results shown that the MolAttFP is essential to get better results than DMPNN. So I decided to check what is the impact of MolAttFP part in top of native GAT called Native+ and for RODE called RODE+. Basically the combination provide an even better RMSE with again an 6.2% improvement versus the Native+ version. It is interesting to observed that the +MolAttFP delivers the most stable results and reaches an 0.55 RMSE.

Figure 19 : Number of Steps for SCORE-GAT versus GAT with MolAttFP option, dt = 1/N

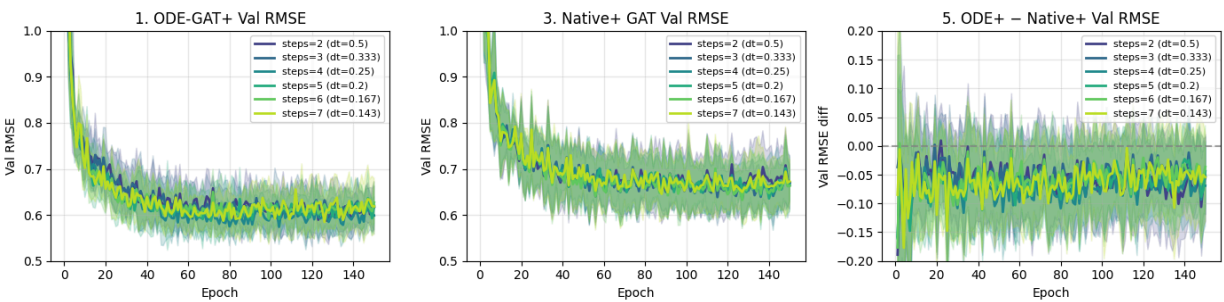

Validation RMSE comparison, by changing number of steps (dt = 1/N) from 2 to 7, SCORE-GAT + MolAttFP and GAT + MolAttFP

Table 7 : SCORE-GAT vs GAT with MolAttFP at best validation RMSE

| steps | Δt | SCORE+ | Native+ | Diff | Improvement |
|---|---|---|---|---|---|
| 2 | 0.5 | 0.559 | 0.618 | +0.059 | 9.6% |
| 3 | 0.334 | 0.566 | 0.598 | +0.032 | 5.3% |
| 4 | 0.25 | **0.545** | 0.610 | +0.065 | 10.6% |
| 5 | 0.2 | 0.564 | 0.610 | +0.045 | 7.5% |
| 6 | 0.167 | 0.564 | **0.602** | +0.038 | 6.2% |
| 7 | 0.143 | 0.558 | 0.606 | +0.048 | 8.0% |
| Average | N/A | 0.559 | 0.607 | +0.048 | 7.9% |

5-fold CV average of best Validation RMSE over 150 epochs with MolAttFP no SCORE-MLP

## Study of number of steps / layers in GATv2 + MolAttFP 2 to 7

One well known alternative to GAT is GATv2, an enhanced version that is expensive to compute as the v2 version needs to compute a non linear attentive equation compared to the GAT initial version. We did not see a significant difference between SCORE+GAT versus SCORE+GATv2 or between native+ versions. So we do not need to use the GATv2 version to get the best 0.55 RMSE performance.

Table 8 : SCORE-GATv2 vs GATv2 with MolAttFP at best validation RMSE

| steps | Δt | SCORE+ | Native+ | Diff | Improvement |
|---|---|---|---|---|---|
| 2 | 0.5 | **0.562** | **0.593** | +0.030 | 5.1% |
| 3 | 0.334 | **0.558** | **0.597** | +0.040 | 6.7% |
| 4 | 0.25 | **0.569** | **0.608** | +0.040 | 6.5% |
| 5 | 0.2 | **0.561** | **0.589** | +0.028 | 4.7% |
| 6 | 0.167 | **0.559** | **0.602** | +0.043 | 7.2% |
| 7 | 0.143 | **0.561** | **0.599** | +0.038 | 6.3% |
| **Average** | N/A | **0.562** | **0.598** | +0.036 | 6.1% |

5-fold CV average of best Validation RMSE over 150 epochs

Figure 20 : Number of Steps for SCORE-GATv2 versus GATv2 with MolAttFP option, dt = 1/N

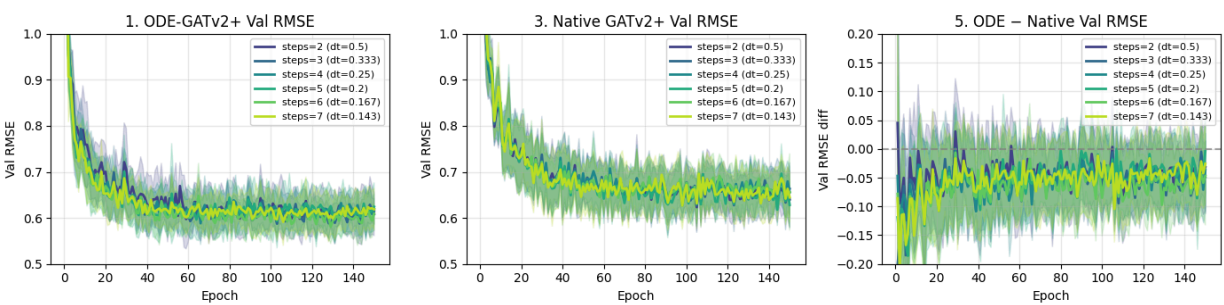

Validation RMSE comparison, by changing number of steps (dt = 1/N) from 2 to 7, SCORE-GATv2 + MolAttFP and GATv2 + MolAttFP

## Study of number of steps / layers in DMPNN + MolAttFP 2 to 7

We decided to also run DMPNN with the additional MolAttFP trick. And the result was very great, showing that the native+ model is very robust and that the ODE+ version still get improvement even if we are touching the limit of the data noise (see CatBoost results).

Table 9 : SCORE-DMPNN vs DMPNN with MolAttFP at best validation RMSE

| steps | Δt | SCORE+ | Native+ | Diff | Improvement |
|---|---|---|---|---|---|
| 2 | 0.5 | **0.548** | 0.574 | +0.026 | 4.6% |
| 3 | 0.334 | 0.553 | **0.551** | -0.002 | -0.4% |
| 4 | 0.25 | **0.543** | 0.562 | +0.019 | 3.4% |
| 5 | 0.2 | **0.557** | 0.560 | +0.003 | 0.5% |
| 6 | 0.167 | 0.564 | 0.560 | -0.004 | -0.6% |
| 7 | 0.143 | **0.552** | **0.555** | +0.004 | 0.7% |
| **Average** | N/A | 0.553 | 0.561 | +0.008 | 1.4% |

5-fold CV average of best Validation RMSE over 150 epochs

Figure 21 : Number of Steps for SCORE-DMPNN versus DMPNN with MolAttFP option, dt = 1/N

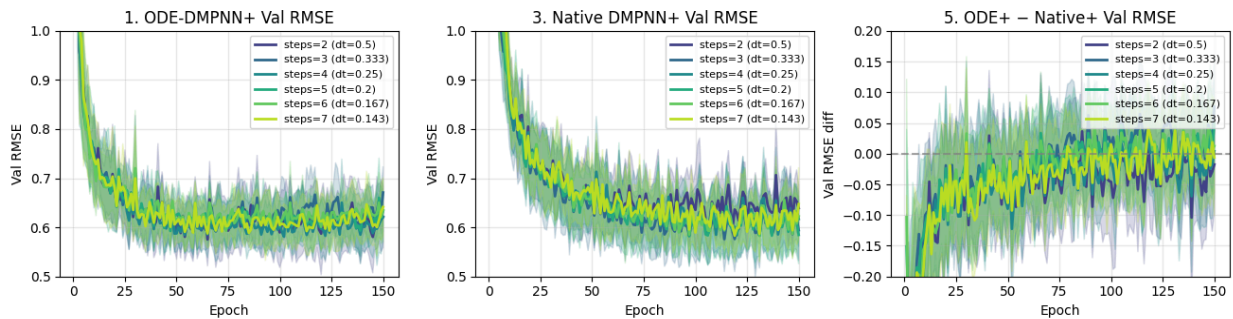

Validation RMSE comparison, by changing number of steps (dt = 1/N) from 2 to 7, SCORE-DMPNN + MolAttFP and DMPNN + MolAttFP

## Learnable Δt

Very recently a learnable skip connection for Graph Neural Network convolution called Adaptive Initial Residual Connection was proposed(29). I proposed to setup the learnable Δt [0.1,0.5] parameter using this equation Δt = 0.1 + 0.4 * σ(α) via the sigmoid function to constraint the system. The goal is to make the system make a dynamic gate to combine the current and previous knowledge using equation 2. Current results did not show any benefits so far as I got very similar results as the Δt preset by number of steps.

## nanoGTP using SCORE

I have tested the Muon Optimizer in order to compress the default nanoGPT version using simple characters tokenizer on a tiny Shakespeare dataset. This task is more complex for the SCORE method than the Native nanoGPT variant at 11 M parameters for 3000 iterations. I observed that SCORE needed a very small dropout and preferred Adam to AdamW as we need to leverage all the weights. Muon allowed the convergence better than Adam on a m_SCORE version 3.6M parameter, consisting of 2 stacking SCORE blocks (M = 2, steps = 3). The model training takes the same time to reach 1.57 val loss, almost the same as the original nanoGPT 11M with AdamW at 1.56 val loss. This is really nice to see that the Muon has this capability to leverage all the weights of the matrix instead of the sparse idea used by default in LLM. Basically this shows the fact that the Optimizer is essential for SCORE SLM models. The SCORE method (steps = 6) with 1.8M parameters also reaches 1.56 val loss. In table 10, we generate example sentences based on training with Muon optimizer of 10.7 M, 3.6 M and 1.8 M models while the original nanoGPT reaches 1.56 with AdamW optimizer, it only achieves 1.6 with Muon (with dropout 0.2 or 0.01) with learning rate of 3e-4.

Table 10 : SCORE, mSCORE and nanoGPT generator examples

| nanoGPT : 10.7 M, 6 layers<br>val loss 1.57 | mSCORE : 3.6 M 2 x 3 steps<br>val loss 1.56 | SCORE : 1.8 M , 6 steps<br>val loss 1.56 |
|---|---|---|
| Is Head you such again; for that ever<br>Wherein we wear you have gover father.<br>LUCIO:<br>My conscient, gopinion to and hopece<br>Insued bewing grions himself spardon'd:<br>To will you will be more. Marcius, which, and we curse at<br>partieng he manague in' hands by the swom:<br>And before law with such rems; her and their her<br>praye love a bearer hein<br>story-merris;<br>And he mans giver the humornt another!<br>This the farful king's tearsure for his loving.<br>JULIET:<br>With thy contrary, may armshal, and neight didined,<br>L | Is that lady for her have set a detter, grace!<br>SICINIUS:<br>The fly sovereignce what thou canst she do?<br>Secrvivant;<br>Boundstand there, not he nobld!<br>LEONTES:<br>Yet ask will'd to do,<br>Signolous I have slel:<br>I fear the way arreate they do good grief,<br>The abjoure sweetss; know when I will stay re abundred<br>Int clean my hand love a bear they could did<br>These such revengealess of any tritle over<br>Tyruesther's friends ame thank a me do them;<br>The pudge in shrow!--O contemn like throw his an;<br>Test mude what of | I'll askly the die.<br>Shepherd:<br>Let thee grace with and what I do not late?<br>Is Lord the loss, where isparish thy richzance<br>In mades with griefs have can pardon'd:<br>To will you with a good consul, like love,<br>And because and any they do power are mine.<br>LEONTES:<br>What, fellow, what were strive, and true?<br>A clown:<br>And he love a be?<br>CAPULET:<br>IsABELLA:<br>Ratueous man'st torment him, it am slaved to the life it know;<br>She am sade he light, being and noble sir:<br>Show do many of requests |

By adding Bigram hash embedding(30), NorMuon with Polar express(31), and replacing LayerNorm by RMSnorm, we can reach with 12 steps 1.53 and best 1.51. Bigram is really accelerating a lot in the early training steps while NorMuon(32) and RMSNorm(33) are fine tuned versions to be more precise and have less parameters. Here is a final prompt generator score model with 1.8 M with 12 steps train with 90/10 or 10/90 train-validation split (the validation loss diverges a lot in this case as we do use only 10% of data for training and we cannot compare with the 1.51 validation loss anymore). But we can see that the Generated texts have both the speakspare style in example from table 11.

Table 11 : SCORE generator examples train with 10/90 or 90/10 splits

| Generated text with 10/90 train-val split | Generated text with 90/10 train-val split |
|---|---|

| best val loss | best val loss 1.51 |
|---|---|
| ICINIUS:<br>May from thence:<br>Lest have you to cry him:<br>Indeed for your cosuntry.<br>BRUTUS:<br>'Tis most like he words his house wints,<br>As he would shall be form all your op?<br>CORIOLANUS:<br>What then shall I bloods I dishonours hip<br>That wouldst do thee with work me invoits.<br>SICINIUS:<br>The crose of crue at at dat the no midnifit<br>Ag the breation walls, but I amoure ompet<br>That thou speak'd it follow.<br>First Senator:<br>No, Caius Caius Marcius coming to Marcius.<br>All:<br>MARCIUS:<br>The deare i' the pears uside the f | I'll ask the fool.<br>LUCIO:<br>Let me to her be you had<br>May the firew's grace less! all she shall set himpance.<br>Shall; how, sir, my king dear love hath can<br>Savance you to knee to do it.<br>CAPULET:<br>Oh, sir, go and with way all and the ground<br>To reap him was foreful nose; kneel not we creature<br>From us ready that there is follow a bed?<br>Came couldst me stay.<br>MAMILLIUS:<br>O grace!<br>And with mostre!<br>BUCKINGHAM:<br>Her sake, I do think me to look as to-morrow<br>Asside thou deadly by arms? Obear himself?<br>To thy of |

Table 12 : Advanced Autosearch on M3 Max 128 GB (5 min is representing the true constraint hardware constraint)

| Config using stable NorMuon | val_bpb | Params | Trials |
|---|---|---|---|
| d4 + skip05 (yours) | 1.2594 | 22M | 1 manual |
| d4 no skip05 (yours) | 1.2621 | 22M | 1 manual |
| SCORE 2-recursive (yours) | 1.2731 | 18.4M | 1 manual |
| PR#4 d4 (BL3IP) | 1.2809 | 22M | 110 via autosearch |